\documentclass{article}

\usepackage{spconf,amsmath,graphicx,csvsimple, siunitx,float,amssymb,nccmath}

\title{Two-branch multi-scale deep neural network for generalized document recapture attack detection}
\name{Jiaxing Li, Chenqi Kong, Shiqi Wang, and Haoliang Li\thanks{This work was supported in part by CityU New Research Initiatives/Infrastructure Support from Central (APRC 9610528), and the Research Grant Council (RGC) of Hong Kong through Early Career Scheme (ECS) under the Grant 21200522.}}
\address{City University of Hong Kong}
\begin{document}
%
\maketitle
\begin{abstract}
The image recapture attack is an effective image manipulation method to erase certain forensic traces, and when targeting on personal document images, it poses a great threat to the security of e-commerce and other web applications. Considering the current learning-based methods suffer from serious overfitting problem, in this paper, we propose a novel two-branch deep neural network by mining better generalized recapture artifacts with a designed frequency filter bank and multi-scale cross-attention fusion module. In the extensive experiment, we show that our method can achieve better generalization capability compared with state-of-the-art techniques on different scenarios. 

\end{abstract}
%
%
\section{Introduction}
\label{sec:intro}
With the growing application of digital images, there is great demand for reliable detection algorithms against different image manipulation methods, since a successful attack on images containing personal confidential information (such as document images) will pose great threats to property safety and privacy. Currently, the direct modifications on the original image file is well encountered by analyzing file-based forensic traces, such as EXIF format and content [1, 2], and the traces of multiple compressions [3, 4], however, by recapturing the distorted images using another device such forensic traces can be effectively erased, thus increasing the manipulated images’ authenticity [5].

\begin{figure}[t]
    \centering
    \includegraphics[width=0.85\linewidth]{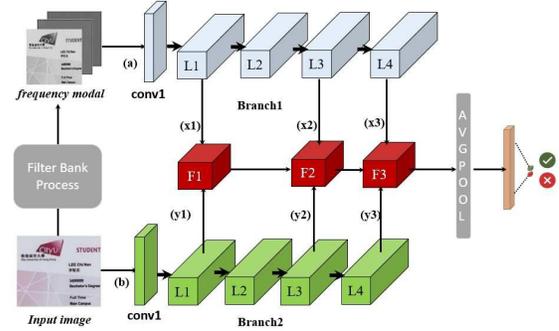}
    \caption{Pipeline of the proposed architecture.}
    \vspace{-10pt}
\end{figure}

The detection methods for recapture attacks can be generally divided into two categories, hand-crafted features-based methods, and learning-based methods. Early methods implemented support vector machine [6] based on the different hand-crafted features such as Local Binary Pattern (LBP) [7], multi-scale wavelet decomposition statistics [8], and Markov-based features [9, 10]. However, these methods have limited accuracy as Agarwal's \textit{et al.}[11] study showed their poor performance when facing a large sophisticated dataset. Heading into the era of artificial intelligence, data-driven methods like convolution neural networks (CNN) [11] with different architectures, and central difference CNN [12] outperformed the hand-crafted-feature-based methods. But these data-driven methods suffer from the lack of generalization capability to unseen domains. As discussed by Chen \textit{et al.}[13], the attacker can implement multiple different device combinations in the steps of capture and recapture, domain shift can be created by using unseen device combinations from the training dataset, and the current generic CNN usually behaved poorly when facing this kind of domain cross situations. By training a siamese network to compare the questioned images with both genuine and recapture reference samples Chen's approach [13] outperformed most of generic CNNs on the device domain cross situation. However, it required an already existing reference pair that has the same content template as the questioned sample to perform examination, which is a major limitation in deployment, since factors like different versions and institutions may result in a lack of same template references. Moreover, in a real detection scenario, the detector also needs to face the domain shift created by different image qualities, for example, the model trained on high-quality images may face test samples that were JPEG compressed.

In this paper, to tackle the domain shifts in recapture attack, we propose a novel algorithm to learn the discriminative artifacts introduced by the recapturing step. Inspired by the previous research works that the recapture process will cause the loss of fine details which mainly located in images’ high frequency domain, and also the color distortion, thus we propose a novel robust two-branch deep neural network where the first branch of the model contains a frequency filter bank preprocess module to better extract the detail loss artifacts, and in the second branch the color distortion artifacts are learned from the original RGB input, and the learned features of the two branches will be aligned in different level through a proposed multi-scale cross-attention fusion module to generate the final prediction. The experiment results show our proposed method obtains state-of-the-art performance towards different domain cross scenarios, and in the ablation study we further prove the reasonability and effectiveness of proposed architecture.

\section{METHODS}
\subsection{Overview}
\noindent Fig.1 illustrates the complete pipeline of our framework. Given an input image, the detail loss artifacts are extracted by the first branch (branch1 in Fig.1) with a filter bank preprocess module, and the color distortion artifacts are mined by the second branch (branch2 in Fig.1) from RGB input, the final predictions are generated by fusing discriminative features mined from frequency and RGB modals. Our proposed model can be trained in an end-to-end fashion with cross-entropy loss. The details of the two key components of our proposed model, input data preprocess, and multi-scale cross-attention fusion process will be introduced below.

\subsection{Input Image Processing}
\textbf{Filter Bank Preprocess Module.} Previous research [5] has shown the recapture process will cause the loss of fine details due to the mismatch between screen and camera resolution, and Cao \textit{et al.} [5] proposed to extract these artifacts from images' high-frequency information using the multi-wavelet decomposition. Later, Li \textit{et al.} [14] implemented a learnable preprocessing filter where the training result showed its spectrum response contained a high pass filter, which also indicated the discriminative features' existence in the high-frequency band.
Based on these findings, we design a filter bank composed of three bandpass filters to extract input images' band-wise information and compose them into a new modality. The complete converting process is shown in Fig.2, First, the RGB input image will be converted into grayscale. Second, the grayscale image will be transformed into the frequency domain through a two-dimensional discrete cosine transform (DCT) [15], and the output is a two-dimensional matrix composed of DCT coefficients. Third, we construct three different bandpass filters. Equation below demonstrates the creation of a low bandpass filter as an example,



\vspace{0.2cm}
\begin{equation}
\abovedisplayshortskip=-10pt
\scriptstyle
B_{low}(i, j)=
\begin{cases}
    $1$ & \text{if $\,0\leq(i+j) < k$},\\
    $0$ & \text{if $\,k\leq (i+j) < 2\cdot k$ },\\
    $0$ & \text{if $\,2\cdot k \leq (i+j)$ },
\end{cases}
\end{equation}
\vspace{-0.1cm}
\noindent where $i, j\in(1, 224)$ indicate element location indexes and $k$ is the threshold value to divide the frequency band into low, middle, and high levels. $B_{low}(i, j)$ denotes the low filter's element value at location $(i, j)$, and it's set to 1 when satisfying the first term. Similarly, the $B_{mid}, B_{high}$ filters can be created by setting the filter element's value to 1 when satisfying the second and third terms, respectively. Then the frequency matrix performs element-wise multiplication with three different filters, each of the output matrices will contain DCT coefficients from one frequency band. In the last step, the three different result frequency matrices will be converted back to spatial domain respectively through the inverse discrete cosine transformation. 




Then the three different output grayscale images are concatenated in the order of low, middle, and high. Different from previous approaches, instead of only retrieving limited high-frequency information like [11] and [5], our method contains lossless frequency information of the input image divided by frequency band.

\begin{figure}[t]
    \includegraphics[width= \linewidth]{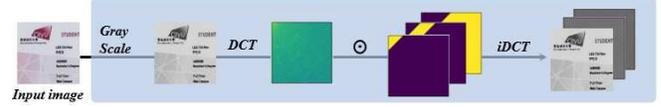}
    \caption{The Complete Frequency Filter Bank Preprocessing.}
\end{figure}

\noindent \textbf{Complementary RGB Image Input.} In [5], it has been shown that the recapture process will also introduce a color distortion, affecting the correlation between different color channels, which can be treated as another discriminative artifact. To explore the color distortion artifact for recapture detection, we propose to use the image in RGB color space as the input of branch2. 



\subsection{Multi-scale Cross-attention Fusion}
\textbf{Cross-Attention Module.} Attention-based fusion methods have shown great success in multi-modal learning tasks such as image-text analysis [16], and it was also applied in computer vision task [17] to capture the dependency between images' different frequency bands. 
Motivated by the success of the cross-attention module, we propose to introduce a novel multi-scale cross-attention module (based on [18]) to capture the correlation between the discriminative forensic features learned from two different branches in different scales, where the architecture of the cross-attention module is shown in Fig.3. The cross-attention process with the output from branch1 can be formulated by the following equation given as
\begin{equation}
\abovedisplayshortskip= -20pt
\belowdisplayshortskip= 25pt
\small
     X_i^a = Softmax[(X_i\star W_i^q)\times(Y_i\star W_i^k)]\times (Y_i\star W_i^v) + X_i,
\end{equation}
where $\star, \times$ indicate convolution operation and matrix multiplication, respectively, $i  \in {\{1, 2, 3\}}$ is the scale index, $X_i, Y_i$ indicate the features extracted by branch1 and branch2 at $i$th scale, $X_i^a$ indicates the output of cross-attention module of branch1 at $i$th scale, and $W_i^q, W_i^k,W_i^v$ represent the corresponding weights for query, key, and value in cross-attention module. The cross-attention computing of the output of branch2 $Y_i^a$ can be performed in a similar way. 

\begin{figure}[t]
\begin{minipage}[b]{\linewidth}
  \centering
  \centerline{\includegraphics[width= 0.85\textwidth]{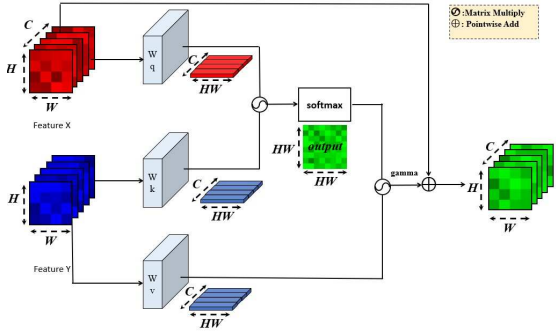}}
  \caption{The cross-attention module.}
\end{minipage}
\vspace{-15pt}
\end{figure}

\noindent \textbf{Multi-scale Fusion.} To fuse the output of cross-attention module at different layers (F1, F2, and F3 shown in Fig.1), we first propose to concatenate the output of branch1 and branch2 followed by downsampling process given as
\begin{equation}
     \scriptstyle
     \begin{aligned}
     A_i = Downsample_{(S_x, S_y)}(Concat(X_i^a, Y_i^a)),
     \end{aligned}
\end{equation}
where $A_i$ indicates the features produced at $i$th scale, and $S_x, S_y$ (where $S_x = S_y = 7$) indicate the output size of downsampling (which is implemented via max pooling and bilinear interpolation). We then fuse the output by concatenating $A_1, A_2$, and $A_3$ in a sequential manner. After concatenating $A_1, A_2$, and $A_3$, we will further employ an average pooling with batch normalization, and forward the output to a MLP module (with 2 layers) to generate the predictions. The whole process is illustrated in Fig.4.

\section{EXPERIMENT}
\label{sec:typestyle}
\subsection{Experiment Setup}
\textbf{Dataset. }Our experiment is conducted on the dataset provided by Chen \textit{et al.}[13], a document image database, it contains Dataset1 (84 genuine images, 588 recaptured images) and Dataset2 (48 genuine images, 384 recaptured images), the two datasets were collected by different devices, and all images were saved in TIFF format. On its base we also create JPEG compressed duplicate datasets for the 2 datasets to test the quality cross performance. 
Images in these two datasets contain 12 different documents, to ensure that the training dataset doesn't contain pre-knowledge about validating and testing datasets, we divide the database according to documents instead of images. And all models are trained on the high-quality dataset1, then we will evaluate the algorithms on dataset2 (cross-dataset evaluation), dataset1 with JPEG image compression (cross-quality evaluation), and dataset2 with JPEG image compression (cross-dataset and cross-quality evaluation)


\begin{figure}[t]
\centering
\includegraphics[width=.55\linewidth]{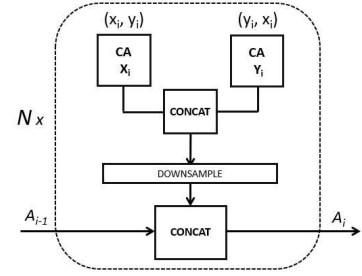}
\caption{The multi-scale fusion mechanism. $N$ denotes the number of multi-scale fusion. We set $N=3$ in our work.}
\vspace{-15pt}
\end{figure}

\noindent \textbf{Baseline Models and Training Setting. }Our proposed architecture is compared with 4 widely used generic CNN architectures with different layer configurations: resnet34, resnet50, resnet152 [19], densenet121, densenet169, densenet201 [20], resnext50 [21] and efficientb4 [22]. All the generic CNNs' classifier is replaced with 2 fully connected layers, layer1 has 256 nodes and layer2 has 2 nodes to adapt the binary classification task. Central difference convolution architecture [12] and siamese network [13] are also compared with our method. We use the resnet50 as the backbone of our proposed model, as it can achieve a desired performance based on intra domain evaluation with relatively small model size. 


In the training process, the $k$ value of the proposed filter bank is set to 10, and all models are trained with input resolution $224 \times 224$, Using Adam optimizer and cross-entropy loss, the batch size is 64, the training epochs is 20, and the learning rate is $10^{-4}$. Our implementation is based on PyTorch 1.12.0.



\vspace{-0.6cm}
\begin{table}[H]
  \abovedisplayshortskip= 10pt
  \belowdisplayshortskip= 10pt
  \caption{Cross-dataset evaluation.}
  \label{cross_mm}
  \centering
  \renewcommand\arraystretch{1.15}
  \scalebox{0.6}{\begin{tabular}{|c|c|c|c|c|c|}
    \hline
    Model & ACC(\%) $\uparrow$ & AUC(\%) $\uparrow$ & EER(\%) $\downarrow$ & AP(\%) $\uparrow$ & HTER(\%) $\downarrow$ \\
    \hline
    \hline
    densenet121	& 72.01 & 80.00 & 27.87	&88.82	&27.07\\
    efficientnetb4	&81.82	&89.90	&18.69	&93.27	&21.50\\
    resnet50	&68.52	&80.10	&24.58	&88.22	&26.70\\
    densenet169	&69.12	&83.92	&22.89	&91.22	&25.28\\
    densenet201	&72.20	&87.44	&20.74	&92.39	&23.77\\
    resnet101	&64.06	&74.30	&30.23	&86.91	&29.75\\
    resnet152	&72.58	&90.86	&15.95	&93.07	&23.02\\
    resnet34	&71.57	&80.07	&27.28	&89.90	&25.24\\
    resnext50	&72.63	&81.23	&25.58	&88.34	&23.55\\
    cdc network[12]	&77.52	&85.90	&21.33	&91.06	&25.19\\
    siamese network[13] &82.20 &87.34 &17.59 &89.92 &17.60\\
    branch1	&79.84	&91.51	&15.33	&94.87	&16.31\\
    proposed(3scale)	&\textbf{84.65}	&\textbf{94.12}	&\textbf{13.29}	&\textbf{96.82}	&\textbf{13.31}\\
    \hline
\end{tabular}}
\end{table}
\vspace{-0.6cm}
\begin{table}[H]
  \abovedisplayshortskip=-20pt
  \belowdisplayshortskip= 10pt
  \caption{Cross-quality (JPEG compression) evaluation.}
  \label{cross_mm}
  \centering
  \renewcommand\arraystretch{1.15}
  \scalebox{0.6}{\begin{tabular}{|c|c|c|c|c|c|}
    \hline
    Model & ACC(\%) $\uparrow$ & AUC(\%) $\uparrow$ & EER(\%) $\downarrow$ & AP(\%) $\uparrow$ & HTER(\%) $\downarrow$ \\
    \hline
    \hline
    densenet121	&89.98	&98.25	&4.26	&99.54	&38.74\\
    efficientnetb4&	90.71	&96.28	&5.88	&99.00	&35.24\\
    resnet50	&92.20	&94.14	&13.88	&99.02	&24.80\\
    densenet169	&92.11	&93.86	&10.01	&98.29	&30.01\\
    densenet201	&94.54	&98.85	&2.84	&99.70	&21.05\\
    resnet101	&95.14	&99.13	&4.11	&99.82	&18.56\\
    resnet152	&92.28	&92.90	&11.40	&98.29	&22.19\\
    resnet34	&93.59	&95.38	&8.25	&98.66	&24.76\\
    resnext50	&91.59	&83.26	&24.81	&95.29	&32.51\\
    cdc network[12]	&90.29	&96.25	&11.62	&99.42	&37.05\\
    siamese network[13]	&83.48	&89.19 &15.52 &87.42 &16.57\\
    branch1	&94.32	&95.18	&9.02	&98.77	&19.24\\
    proposed(3scale)	&\textbf{97.36}	&\textbf{99.76}	&\textbf{1.65}	&\textbf{99.97}	&\textbf{9.23}\\
    \hline
\end{tabular}}
\end{table}

\vspace{-0.6cm}
\begin{table}[H]
  \abovedisplayshortskip=-20pt
  \belowdisplayshortskip= 10pt
  \caption{Cross-dataset and cross-quality evaluation.}
  \label{cross_mm}
  \centering
  \renewcommand\arraystretch{1.15}
  \scalebox{0.6}{\begin{tabular}{|c|c|c|c|c|c|}
    \hline
    Model & ACC(\%) $\uparrow$ & AUC(\%) $\uparrow$ & EER(\%) $\downarrow$ & AP(\%) $\uparrow$ & HTER(\%) $\downarrow$ \\
    \hline
    \hline
    densenet121	&68.45	&80.16	&27.80	&87.81	&41.26\\
    efficientnetb4	&71.16	&81.97	&23.52	&84.52	&39.13\\
    resnet50	&76.75	&81.38	&22.52	&87.55	&22.66\\
    densenet169	&77.69	&86.86	&18.54	&91.55	&25.26\\
    densenet201	&73.06	&84.09	&26.28	&90.47	&29.03\\
    resnet101	&72.21	&77.54	&26.87	&88.63	&24.21\\
    resnet152	&75.88	&87.32	&19.70	&89.41	&21.88\\
    resnet34	&71.57	&80.07	&27.28	&87.09	&26.94\\
    resnext50	&69.88	&78.00	&27.41	&84.11	&34.51\\
    cdc network[12]	&74.46	&83.65	&24.89	&90.05	&31.57\\
    siamese network[13] &72.27	&72.21 &34.87 &79.60 &30.82\\
    branch1	&76.41	&86.37	&21.19	&88.48	&32.07\\
    proposed(3scale)	&\textbf{86.16}	&\textbf{93.37}	&\textbf{13.50}	&\textbf{96.62}	&\textbf{14.89}\\
    \hline
\end{tabular}}
\end{table}

\subsection{Results}
We first study the effectiveness of our proposed filer bank preprocess module by comparing using branch1 of our model with the generic resnet50 using RGB image as input. As we can observe from Table 1-3, the resnet50 with frequency modal input outperforms RGB input one by a large increase in all domain cross situations, in the cross-dataset scenario, branch1 achieved $11.32\%$ improvements in terms of ACC and $11.41\%$ improvements in terms of AUC, and in the cross-quality scenario, branch1 obtained $2.12\%, 1.04\%$ improvements in ACC, AUC respectively, which shows that the fine details loss artifact can be effectively extracted through the filter bank preprocess. And during the cross-dataset and cross-quality situation, branch1 obtains a $4.99\%$ improvement in terms of AUC and achieves competitive performance in terms of ACC compared with the generic resnet50. The results further justify the effectiveness of our adopted filter bank processing.

We can also observe that our proposed multi-scale architecture proposed(3scale) achieves state-of-the-art performance in all situations. In the cross-dataset scenario, our proposed method outperforms the Siamese network [13] method by $2.45\%$, and $6.78\%$ in terms of ACC and AUC respectively (as shown in Table 1). Regarding the cross-quality scenarios,
 the proposed method obtains $97.36\%$ accuracy and $99.76\%$ AUC score which outperform the baseline based on central difference convolution [12] by $7.07\%$ and $3.51\%$ (as shown in Table 2). In the cross-dataset and cross-quality, the proposed method outperforms the second-best method, generic CNN densenet169, by $6.51\%$ in terms of AUC (as shown in Table 3). The result demonstrates the effectiveness of our algorithm towards recapture attack detection by learning both fine detail loss and color distortion artifacts introduced by recapturing.
  
\subsection{Ablation Study}
We first study the effectiveness of the RGB complement branch, by comparing branch1 with proposed(3scale) the complete framework we can see significant improvement can be achieved based on all scenarios,  which shows that branch2 can help to extract color-related information, which benefits document recapture attack detection. 


\begin{table}[H]
  \abovedisplayshortskip=-20pt
  \belowdisplayshortskip= 10pt
  \caption{Ablation study on cross-dataset evaluation. }
  \label{cross_mm}
  \centering
  \renewcommand\arraystretch{1.15}
  \scalebox{0.6}{\begin{tabular}{|c|c|c|c|c|c|}
    \hline
    Model & ACC(\%) $\uparrow$ & AUC(\%) $\uparrow$ & EER(\%) $\downarrow$ & AP(\%) $\uparrow$ & HTER(\%) $\downarrow$ \\
    \hline
    \hline
    branch1	&79.84	&91.51	&15.33	&94.87	&16.31\\
    base-fusion	&72.21	&88.09	&19.64	&89.91	&25.42\\
    proposed(1scale) &70.15	&83.87	&23.35	&90.58	&24.92\\
    proposed(2scale)	&80.62	&93.38	&14.82	&96.11	&16.39\\
    proposed(3scale)	&\textbf{84.65}	&\textbf{94.12}	&\textbf{13.29}	&\textbf{96.82}	&\textbf{13.31}\\
    \hline
\end{tabular}}
\end{table}

\vspace{-0.6cm}
\begin{table}[H]
  \abovedisplayshortskip=-20pt
  \belowdisplayshortskip= 10pt
  \caption{Ablation study on cross-quality evaluation.}
  \label{cross_mm}
  \centering
  \renewcommand\arraystretch{1.15}
  \scalebox{0.6}{\begin{tabular}{|c|c|c|c|c|c|}
    \hline
    Model & ACC(\%) $\uparrow$ & AUC(\%) $\uparrow$ & EER(\%) $\downarrow$ & AP(\%) $\uparrow$ & HTER(\%) $\downarrow$ \\
    \hline
    \hline
    branch1	&94.32	&95.18	&9.02	&98.77	&19.24\\
    base-fusion	&93.45	&94.87	&9.99	&98.85	&28.32\\
    proposed(1scale)	&94.79	&98.68	&5.38	&99.73	&20.12\\
    proposed(2scale)	&95.31	&99.72	&2.05	&99.96	&17.80\\
    proposed(3scale)	&\textbf{97.36}	&\textbf{99.76}	&\textbf{1.65}	&\textbf{99.97}	&\textbf{9.23}\\
    \hline
\end{tabular}}
\end{table}

\vspace{-0.6cm}
\begin{table}[H]
  \abovedisplayshortskip=-20pt
  \belowdisplayshortskip= 10pt
  \caption{Ablation study on cross-dataset and cross-quality evaluation.}
  \label{cross_mm}
  \centering
  \renewcommand\arraystretch{1.15}
  \scalebox{0.6}{\begin{tabular}{|c|c|c|c|c|c|}
    \hline
    Model & ACC(\%) $\uparrow$ & AUC(\%) $\uparrow$ & EER(\%) $\downarrow$ & AP(\%) $\uparrow$ & HTER(\%) $\downarrow$ \\
    \hline
    \hline
    branch1	&76.41	&86.37	&21.19	&88.48	&32.07\\
    base-fusion	&77.75	&84.57	&21.95	&91.73	&25.42\\
    proposed(1scale)	&78.26	&84.75	&22.45	&90.99	&21.12\\
    proposed(2scale)	&85.08	&92.49	&15.41	&95.65	&16.11\\
    proposed(3scale)	&\textbf{86.16}	&\textbf{93.37}	&\textbf{13.50}	&\textbf{96.62}	&\textbf{14.89}\\
    \hline
\end{tabular}}
\end{table}

We then evaluate the effectiveness of our proposed method of cross-attention fusion by comparing with the baseline without cross-attention module (base-fusion in the tables, where we only perform feature concatenation together with downsampling process). As we can see, without our cross-attention module, the performances also drop significantly in all scenarios. The results further justify the superiority of our introduced cross-attention module. 


Last but not the least, We study the effectiveness of our proposed multi-scale configurations by comparing proposed(3scale) with proposed(2scale) (where F1 module in Fig.1 is removed) and proposed(1scale) (where F1, F2 module in Fig.1 are removed). Based on the results, we can see that by considering multi-scale information, the optimal performance can be obtained. 
 
\label{ssec:subhead}

\label{sec:pagesityle}

\section{CONCLUSION}
In this paper, we proposed a novel two-branch multi-scale deep neural network utilizing both fine detail loss and color distortion artifacts for document recapture attack detection. Specifically, we extract the detail loss artifact through the designed filter bank process, and a novel two-branch deep neural network is designed in a multi-scale cross-modal-attention fusion mechanism. Experiment results show that the proposed architecture obtains state-of-the-art performance on different types of recapture scenarios. 
\vfill\pagebreak

\bibliographystyle{IEEEbib}
\bibliography{strings,refs}

\end{document}